\pdfoutput=1

\documentclass[11pt]{article}

\usepackage[]{acl}
\usepackage{tabularx}

\usepackage{times}
\usepackage{latexsym}
\usepackage{multirow}
\usepackage{graphicx}
\graphicspath{ {./images/} }
\usepackage{amsmath}
\usepackage{titling}
\usepackage{color,soul}

\usepackage[T1]{fontenc}

\usepackage[utf8]{inputenc}

\usepackage{microtype}

\title{Main Predicate and Their Arguments as Explanation Signals For Intent Classification}

\author{\textbf{Sameer Pimparkhede, Pushpak Bhattacharyya}\\
Indian Institute of Technology Bombay\\
\{sameerp, pb\}@cse.iitb.ac.in}

\date{}  

\begin{document}
\maketitle
\begin{abstract}
Intent classification is crucial for conversational agents (chatbots), and deep learning models perform well in this area. However, little research has been done on the explainability of intent classification due to the absence of suitable benchmark data. Human annotation of explanation signals in text samples is time-consuming and costly. However, from inspection of data on intent classification, we see that, more often than not, the main verb denotes the action, and the direct object indicates the domain of conversation, serving as explanation signals for intent. This observation enables us to hypothesize that {\bf the main predicate in the text utterances along with the arguments of the main predicate can serve as explanation signals}. Leveraging this, we introduce a new technique to automatically augment text samples from intent classification datasets with word-level explanations. We mark main predicates (primarily verbs) and their arguments (dependency relations) as explanation signals in benchmark intent classification datasets ATIS and SNIPS, creating a unique 21k-instance dataset for explainability.
Further, we experiment with deep learning and language models. We observe that models that work well for classification do not perform well in explainability metrics like \emph{plausibility} and \emph{faithfulness}. We also observe that guiding models to focus on explanation signals from our dataset during training improves the \emph{plausibility} Token F1 score by 3-4\%, improving the model's reasoning.
\end{abstract}

\section{Introduction}

\begin{table*}[t]\centering
\begin{tabular}{c c}
 Text & Where can I \hl{watch} the \hl{television} \hl{show} The Private Affairs of Bel Ami ? \\
 Label & SearchCreativeWork \\ 
 \hline
Text & \hl {Book} a \hl{table} at the top-rated \hl{pub} in Garner \\
Label & BookRestaurant \\ \hline
Text & What are my \hl{meal} \hl{options} on airlines from Boston 066
to Denver \\
Label & atis-meal \\ \hline
Text & what \hl{price} is a \hl{limousine} \hl{service} to New York's La Guardian \\
Label & atis-ground-fare
\end{tabular}\\
\caption{Examples of text utterances and ground truth Labels from SNIPS and ATIS datasets. The highlighted part denotes the explanation signals marked using the Main predicate and its arguments.}
\label{Table 1}

\end{table*}

Intent classification is widely used in real-life chatbots like Alexa, Siri, and other conversational AI tools. It is used to grasp the nuanced aspects of user expressions. It involves (i) comprehending the intentions behind the expressions to guide the agent's actions and (ii) identifying slots that indicate crucial entities necessary for executing the actions. Intent identification is studied as a multiclass classification problem by \citet{raymond2007generative}. However, it is observed that jointly solving it with slot filling improves the performance \citep{qin2021co,chen2019bert}. While deep learning models perform well in intent classification for specific datasets \citep{chen2019bert,gunaratna2022explainable}, they face challenges in generalizing across diverse domains and fine-grained intents, particularly for classes with scarce data \citep{elazar-etal-2021-amnesic,casanueva-etal-2020-efficient}. In scenarios with numerous conversation domains and fine-grained intent classes, acquiring significant data for every class becomes challenging. Additionally, the presence of shared words among fine-grained classes with subtle variations in meaning adds complexity to the task. In the text \emph{Find me an airline with a meal facility}, the intent is \emph{atis-airline} as the focus is on the airline, not the meal. While \emph{meal} is a key entity, it doesn't define intent.
Conversely, in \emph{What are my meal options on airlines from Boston to Denver}, the intent is \emph{atis-meal} as the user asks about meals, not the airline. However, the deep learning model learns from correlations and frequently used words. Hence, it misclassifies the later utterance as atis-airline due to the abundance of that class in the training data and the higher frequency of the term \emph{airline}. This leads to a low overall F1 score and low accuracy for classes with fewer samples (Table \ref{Table 3}). These misclassifications severely affect task-oriented dialogue systems. For instance, the text \emph{At the Charlotte airport, how many different types of aircraft are there for us?} is often misclassified by BERT as atis-airport instead of atis-aircraft due to an imbalance in sample sizes. This can mislead the conversation ahead. However, performance is improved if the model focuses on the right keywords, like aircraft. Due to this, it is essential to analyze the models' reasoning for intent classification and improve their performance.

\begin{figure*}
  \includegraphics[width=\textwidth,height=4cm]{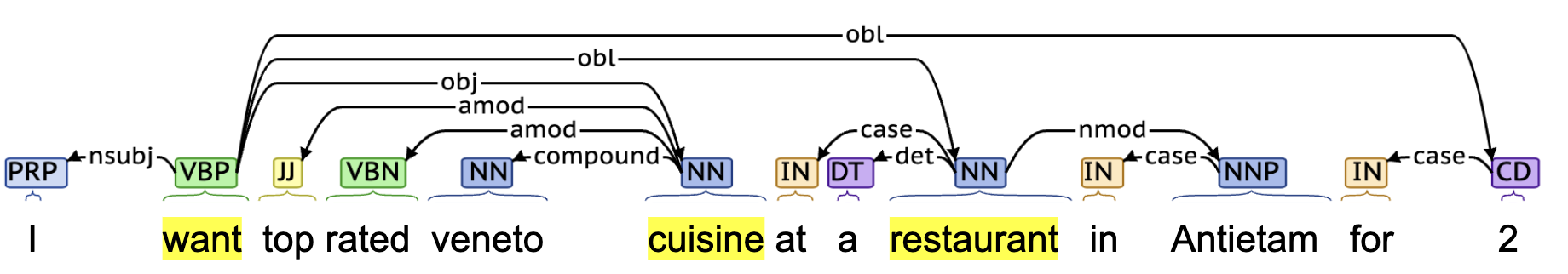}
  \caption{Dependancy parse tree of text sample from SNIPS dataset having a label as BookRestaurants. Highlighted tokens are marked as explanation signals according to the algorithm. This parse tree is fetched using Stanford CoreNLP with OpenIE in the background}.
\label{Figure 1}
\end{figure*}

Unfortunately, even if we get the model's explanation for intent identification, there are no ground truth datasets to evaluate it. Generally, such datasets are created using human annotation, but it is costly and time-consuming \citep{mathew2021hatexplain,deyoung2019eraser,hayati-etal-2021-bert}. Recent methods use large language models (LLM) to generate synthetic explanations. This approach might provide unreliable explanations and, hence, is difficult to integrate into business \cite{li2022explanations, ye2022unreliability}. Hence, we create word-level explanation signals automatically using an interesting linguistic observation. We observe that the main verb denotes action, and its direct object denotes the domain of conversation. Hence, they can act as explanation signals for the intent class. Inspired by these observations, we introduce a novel silver standard annotation technique that uses dependency relations among words as shown in figure \ref{Figure 1} and augments text samples from intent classification datasets with word-level explanation signals. Using this technique, we derive high-quality silver standard explanations for two intent classification datasets ATIS \cite{hemphill-etal-1990-atis} and SNIPS \cite{coucke2018snips}. Hence, we augment 21k text samples with explanation signals and create the first benchmark dataset for explainability in intent classification. We perform the human evaluation of silver annotations to verify the quality of automatically generated explanation signals. For examples from our dataset, refer to table \ref{Table 1}.

Further, we investigate the reasoning of popular deep learning and language models. Explanations are derived from models in the form of saliency maps using standard post-hoc explanation techniques LIME \cite{ribeiro2016should} and integrated gradients \cite{sundararajan2017axiomatic}. We observe that the models that work well for intent classification often do not perform well on explainability metrics \emph{plausability} and \emph{faithfulness} (refer to table \ref{Table 2} and \ref{Table 4}). To improve the reasoning, we guide the model to focus more on silver annotated explanation signals from our dataset during training. It helps the model make more human-like decisions and improves performance for explainability metrics. Our system is ideal for chatbots in small enterprises that can't afford in-house and costly models like GPT-4. Instead, improving reasoning for smaller models like LSTM, BERT, and GPT2, which give efficient responses, can benefit intent classification in chatbots. One can quickly annotate the explanation signals using the main predicate and arguments in the ongoing conversation and guide the deep learning models to focus on those signals more.

\noindent\textbf{Our contributions are}:
\begin{enumerate}
    \item A Novel technique to create silver annotations of word-level explanation signals using the main predicate and its arguments for the text samples from standard intent classification datasets (Figure \ref{Figure 1} and section \ref{sec:algorithm}) with a focus on explainability, which is highly relevant for dialogue agents and chatbots used in small-scale enterprises.
    \item A First-of-its-kind dataset with high-quality word-level explanations of the existing intent classification datasets ATIS and SNIPS consisting of $20k$ samples.
    \item Detailed experiments to investigate the explainability performance in \emph{plausibility} and \emph{faithfulness} of deep learning and language models on our dataset (Table \ref{Table 2} and \ref{Table 4}). We create the first benchmark for explainability in the domain of intent classification.
\end{enumerate}

\section{Related work}

Intent classification is a well-studied problem in natural language understanding (NLU). While intent classification is a sentence-level classification task, slot filling is a more challenging task that deals with classifying the type of each word. These problems can be solved independently \citet{raymond2007generative}, but they are generally solved jointly to optimize performance \citet{chen2019bert,qin2021co}. Our primary focus here is to investigate and improve the model's reasoning for intent classification.

In explainability, post-hoc explanation techniques are popular and useful for feature attribution. This includes perturbation-based methods \citep{ribeiro2016should,ribeiro2018anchors}, Gradient-based methods like integrated gradient \cite{sundararajan2017axiomatic} and attention-based methods \citep{wu2021explaining, wang-etal-2016-attention, pmlr-v37-xuc15}. For natural language processing (NLP) applications, feature importance is measured by the attribution score assigned to every token. These methods explainform of saliency maps, which is a suitable form for NLP due to well-defined features like words and phrases.

Many methods go beyond evaluation and ingest feature attribution priors during training. Some methods use attribution scores derived from post-hoc explanation techniques to train a hate-speech classifier under a scarce data scenario like \citet{liu2019incorporating}. \citet{zhong2019fine} and \citet{mathew2021hatexplain} supervised the model's attention using human-annotated rationale. \citet{jayaram-allaway-2021-human} incorporates feature attribution for documents from the legal domain. However, these methods primarily focus on tasks like hate speech or sentiment classification. We instead explore the area of intent classification.

Regarding explainability in intent classification, \citet{joshi2021towards} uses Layer wise relevance propagation (LRP) \citep{montavon2019layer} to investigate the deep learning model's reasoning over the ATIS dataset. However, this study only evaluates qualitatively based on some examples due to a lack of ground truth explanation signal data. To solve this problem, we introduce a benchmark dataset using the novel silver annotation technique and evaluate the models' reasoning quantitatively based on multiple metrics. \citet{gunaratna2022explainable} focuses on explaining slot classification, not intent.

\section{Dataset}
We use two intent classification benchmark datasets, ATIS \cite{hemphill-etal-1990-atis}, and SNIPS \cite{coucke2018snips}. ATIS consists of a set of text utterances and their corresponding intent labels. It has 4478, 500, and 893 text utterances with gold intent in train, development, and test set, respectively. It consists of 21 intents about the airline travel domain. The second dataset, SNIPS \cite{coucke2018snips}, consists of 13,084 training utterances; the validation and test sets consist of 700 utterances each, with 100 queries per intent. It has seven different intent classes, each belonging to a different domain. Text utterances are from domains like restaurants, movies, music, etc.

We select these datasets with a specific purpose. ATIS has fine-grained intents, and some classes have a low training and test set size (refer to table \ref{Table 3}). This imbalance enables us to evaluate the generalizability and reasoning of the deep learning models when dealing with data scarcity. On the other hand, the SNIPS dataset is balanced but consists of utterances from multiple domains. We can use the SNIPS dataset to evaluate the model's reasoning for intents from diverse domains. Evaluating the reasoning of deep-learning-based models on both aspects gives a good analysis. 

We mark word-level explanation signals in both the ATIS and SNIPS datasets. Hence, we contribute around 21k samples with explanation signals. From ATIS, we filter out the utterances with more than one intent, and we also filter classes in the test set but not in the train set. The average length of silver annotated explanation for the SNIPS dataset is 3.4 words, while it is 4.2 for the ATIS dataset.
\section{Silver annotated explanations}
\label{sec:length}

Human annotation of word-level explanations for intent classification is difficult. Deciding whether original slots in the dataset should be added as an explanation signal is subjective and generally requires multiple human annotators. For example, text utterance, \emph{Show me the flights from Pittsburgh to Los Angeles on Thursday} has gold intent of \emph{atis-flight}. In this example, Pittsburgh and Los Angeles are slots with slot-types source city and destination city. These entities are essential to execute appropriate further actions. However, it does not explain the intent because changing the city names to other cities does not affect the intent of the utterance. Generally, multiple humans annotate explanation signals, which is costly and time-consuming. Hence, we propose a method to introduce word-level explanation signals automatically to existing intent classification datasets consisting of text utterance and gold intent.
We define our dataset as follows. $N$ is the number of samples in the dataset, and each text sample $X_i$ has $m_i$ tokens denoted by $(w_{1}, w_{2}, w_{3},.....w_{m_i})$. We add word-level explanation signals e denoted by $(e_{1}, e_{2}, e_{3},......e_{m_i})$ where each $e_{j} \in \{0,1\}$. $y_i$ is a gold intent label for every $X_i$, directly adopted from the original intent classification datasets. We add these word-level explanation signals for both ATIS and SNIPS datasets.

\subsection{Using slots in dataset}
ATIS and SNIPS datasets consist of slots and their type, which provide specific conversation-related information. Example: Text utterance "which airlines fly from Boston to Washington DC via other cities" has gold intent label as \emph{ATIS-AIRLINE} and slots marked in B-I-O format as "O O O O B-fromloc. city-name O B-toloc. city-name B-toloc. state-code O O O". Hence "Boston", "Washington" and "DC" are the words which are marked as slots. Although these slots provide essential entities to execute the actions, they fail to deliver an \emph{explanation} for the gold intent atis-airline. Slots do not cover words like "airlines." These words are essential to provide an explanation as the dataset has other intent classes like \emph{atis-aircraft} and \emph{atis-flight}, which have very close meanings. Similarly, in the SNIPS dataset, text utterance \emph{Add artist to playlist Epic Gaming has slots marked as "O B-music-item O O B-playlist I-playlist.} The slots in this text sample are \emph{artist epic gaming}, which does not form a sufficient explanation for gold intent \emph{AddPlaylist}. Moreover, intent does not change, even if we replace "epic gaming" with other playlist names. These examples depict that slots in the dataset can't be used as a sufficient and comprehensive explanation.

\subsection{Main Verbs and Arguments}
\label{sec:algorithm}
Creating silver standard explanation signals for intent classification comes from a simple yet significant insight. In many cases, information for classifying the intent of the text utterance is encoded within the main verb and its direct object.
\emph{Example: Set up an alarm for 9 AM tomorrow} has the intent  \emph{set alarm}
and \emph{delete all the alarms} has the intent \emph{delete-alarm}. Here, the word
\emph{alarm} informs that action has to be performed on the alarm, and the words
\emph{set} and \emph{delete}, which are main verbs, inform about a specific action to be performed.
Now, if we change the domain from setting an alarm to setting a calendar, it is the direct object of the main verb that denotes this change in the domain of conversation.
\emph{Example: Set up a meeting for 9 AM tomorrow} has the intent \emph{set calendar}. However, due to the diverse structures of human language, explanation signals might play different syntactic roles, and relying upon just the main verb and direct object may not provide sufficient explanation. Hence, we design a technique that considers multiple linguistic features using the dependency parse tree, automatically deriving high-quality explanations. The algorithm used to derive word-level explanation signals is as follows:
\\

\noindent \textbf{Algorithm}
\begin{enumerate}
\item Find the dependency tree of a given text utterance using the OpenIE platform.
\item Find the main predicate.
\item Travel the dependency parse tree from the root.
\item  Find children of the main predicate with relation direct object, noun- subject, xcomp (open clausal complement), obl (oblique nominal)
\item In case of multiple obl (oblique nominal) for root word, include the obl with part of speech as a common noun.
\item mark these nodes as explanation signals and level 1 traversal ends here.
\item Travel the children of nodes collected in level 1. If the child node has a relation as "compound" and its part of speech is not a proper noun, mark it as an explanation signal.
\item Include all the visited nodes in the explanation
\item Remove stopwords from the explanation segment.
\end{enumerate}
The algorithm starts by default by selecting the main predicate as the explanation signal. The main verb is not mentioned in the text utterances like \emph{Ground Transportation at Baltimore airport}. Hence, We use the Main Predicate instead of the main verb. It is not always necessary that the main verb or main predicate is enough to explain the action. Hence, we also decided to add the xcomp of the main predicate.  
xcomp acts as a secondary predicate when the main predicate does not add much value as an explanation signal. For example, the text utterance from the ATIS dataset, "I need to fly from Atlanta to Denver" has the main predicate predicted by the OpenIE platform as "need". But, it does not add much value as an explanation of intent. Instead, including the word "fly" which is xcomp of the main predicate, confirms the domain of conversation.
We choose to eliminate all the adjective modifiers, nominal modifiers, and oblique nominal with part of speech as proper nouns. These words may provide some relevant information about the conversation but might not be appropriate as an explanation. For example, the text utterance \emph{Please list all cheapest flights on United from Denver to Baltimore} has gold intent \emph{atis-flight}. The word \emph{cheapest} is an adjective that might be essential to provide specific information. Still, even if we change the word "cheapest" to "costliest," it does not change the intent.

Even though the direct object of the main predicate plays an important role, it does not necessarily provide all the information about the domain of conversation. Instead, providing only that explanation signal might be misleading when two intents are from different domains with similar actions. For example, "Add this item to the grocery" suggests the user is talking about a grocery item, and another statement, "Add this item to a playlist," indicates the user is talking about music. Both the words grocery and playlist have an obl dependence on the main predicate here, which must be included to provide a sufficient and comprehensive explanation.
Finally, we consider cases where the main predicate might come in between, and the subject with part of speech nouns is essential.
After including all the explanation signals, we filter the stopwords to reduce noise further. We use Python NLTK library as a source of stopwords \footnote{\url{https://www.nltk.org/}}. Table \ref{Table 1} shows some examples of added explanation signals. Figure \ref{Figure 1} shows an example of producing explanation signals using a dependency parse tree.

\subsection{Dataset Quality Evaluation}

We evaluate the quality of silver explanation signals generated in the form of the main predicate and its arguments. It is a time-consuming process to evaluate all 21k samples. Hence, we select an equal number of samples for every class and perform the human evaluation for this subset. We provided human evaluators with only the silver annotated explanation segment instead of the complete text. We asked them to select an appropriate intent class, for example, instead of a full-text sample \emph{Where can I watch the television show The Private Affairs of Bel Ami ?}. We provide evaluators only the silver annotated explanation \emph{watch television show} and ask them to select the most appropriate intent label. Evaluators were provided brief definitions of intent classes along with some examples. Evaluation is done separately for ATIS and SNIPS datasets. We randomly selected 200 samples from both datasets. In the selection process, we ensure no class imbalance in any datasets. \textbf{Each sample is evaluated by three evaluators.}
All the evaluators are experts in linguistics and language. Two annotators are Ph.D. candidates in linguistics; the third has completed a master's in English. We use three evaluators per sample due to the numerous classes in both datasets. Particularly, ATIS has 21 fine-grained intent classes. Given subtle differences in meanings among some classes from ATIS, relying on one or two annotators may lead to biased or inconsistent evaluation scores. To mitigate subjectivity, three evaluators annotate each sample's label. Inter Annotator Agreement (IAA) was calculated using Fleiss’ Kappa score. \textbf{IAA kappa score obtained is 0.67 for the ATIS dataset and 0.74 for the SNIPS dataset sample.} Fleiss’ Kappa score between 0.61 and 0.8 is considered to be \emph{substantial} agreement \cite{landis1977measurement}. We evaluate human evaluation accuracy against the gold labels from original datasets. \textbf{The average human accuracy obtained is 96.3\% for the ATIS dataset and 97.8\% for the SNIPS dataset.} All the annotators were paid as per industry standards and the suitability of the task.

Although most of the text samples are identified correctly by evaluators, we perform a detailed analysis of the human evaluation. We observe that text utterances marked incorrectly resemble other classes in terms of their ground truth meaning. For example, text utterance \emph{show me all the Lufthansa flights between Philadelphia and Denver} has ground truth intent as \emph{atis-flight}  and derived explanation using tree traversal algorithm is \emph{show Lufthansa flights}. Two out of three evaluators misclassified it as \emph{atis-airline}. Because \emph{Lufthansa} is the name of the airline company, it is difficult to identify the exact class among \emph{atis-airline} and \emph{atis-flight}. Due to minimal distinction in meanings, the most confused pairs of classes in ATIS were \emph{(aits-airline, atis-flight)} and \emph{(atis-flight, atis-flight-time)}. For the SNIPS dataset, \emph{SearchCreativeWork} consists of text utterances related to movies and songs that resemble very closely to classes \emph{SearchScreeningEvent} and \emph{PlayMusic}. For example, text utterance \emph{Find me the song called The Budapest Beacon} has ground truth intent in SNIPS as \emph{SearchCreativeWork} and the derived explanation is \emph{Find song}. All three evaluators marked intent as \emph{PlayMusic} when provided with an explanation part only. For SNIPS, evaluators were most confused with \emph{SearchCreativeWork} and \emph{SearchScreeningEvent} classes. For these text samples, annotating correct intent may not be possible even with full-text utterances. Hence, this depicts that explanations derived from the main verb and its arguments are indeed high quality.

\section{Experiments}

\subsection{Incorporating feature attribution}
It is observed that ingesting feature attributions during training helps to improve the model's reasoning. We investigate if the model's reasoning improves by guiding it to focus on the silver annotated explanation signals. Following the work of \citet{liu2019incorporating}, we use the attribution prior loss, which calculates attribution values using integrated gradients \cite{sundararajan2017axiomatic}. We define $c$ as the total number of intent classes. Each text utterance $X_i$ consists of $m_i$ tokens. Each token is marked with binary explanation signal $t_{i}$. We make slight variations in the original methodology. \citet{liu2019incorporating} uses the attribution using the derivative of the output concerning only one specific class, but following \citet{jayaram-allaway-2021-human}, we sum over the attributions of all the classes as specified in equation \ref{attribution} for regularization and use its average in equation \ref{extra_loss}. We compute the average across output classes as the impact of every neuron, whether positive or negative, is integral to the model's decision \cite{jayaram-allaway-2021-human}.

\begin{equation}
    a_{i} = \sum_{j=1}^{c} a_{i}^j/c
\label{attribution}
\end{equation}

\begin{equation}
    L^{prior}(a,t) = \sum_{i=1}^{n} (a_{i}-t_{i})^2\
\label{extra_loss}
\end{equation}

This attribution loss is added to the regular cross-entropy loss function used for simple classification. 

\begin{equation}
    L^{joint} = L(y,p) + \lambda\sum_{i=1}^{n} (a_{i}-t_{i})^2\
\label{joint loss}
\end{equation}
\begin{equation}
    L(y,p) = \sum_{c}^{C} - y_{c}*log(y_{c})\
\label{C.E loss}
\end{equation}
Unlike \citet{liu2019incorporating}, we do not penalize the model based on a universal token list. Instead, we use local explanation signals as each sample's target values. Hence, the value of $t_{i} \in \{0,1\}$ within each sample $X_i$ serves as its respective set of target values. This is done because a word that is an explanation signal for one sample may not be an explanation signal to another.

\begin{table*}[t]\centering
  \begin{tabular}{c c c c c c c}
    Dataset & Model & Accuracy & Token F1 & IOU F1 & Comprehensiveness & Sufficiency\\ \hline
    \multirow{8}{*}{\textbf{ATIS}}& CNN & 82.32\% & 0.232 & 0.06 &0.38 & \textbf{-0.068}\\
    & CNN Joint & 86.24\% & 0.261 & 0.06 & 0.34 & -0.34\\
    & LSTM & 93.23\% & 0.234 & 0.07 & 0.42 & 0.001\\
    & LSTM Joint & 95.27\% & 0.266 & 0.08 & 0.45 & -0.03\\
    & BERT & 97.2\% & 0.315 & 0.11 & 0.49 & 0.11\\
    & BERT Joint & 97.6\% & 0.346 & 0.12 & 0.45 & 0.06\\
    & RoBERTa & 97.6\% & 0.323 & 0.11 & \textbf{0.49} & 0.15\\
    & RoBERTa Joint & \textbf{97.8\%} & \textbf{0.354} & \textbf{0.12} & 0.44 & 0.09\\
    & GPT-2 & 88.51\% & 0.234 & 0.06 & 0.43 & 0.003\\ 
    & GPT-2 Joint & 91.63\% & 0.267 & 0.09 & 0.46 & -0.06\\\hline
    \multirow{8}{*}{\textbf{SNIPS}}& CNN & 97.14\% & 0.342 & 0.14 & 0.58 & \textbf{-0.075}\\
    & CNN Joint & 97.17\% & 0.366 & 0.14 & 0.52 & -0.022\\
    & LSTM & 97.83\% & 0.347 & 0.14 & 0.54 & 0.004\\
    & LSTM Joint & 97.86\% & 0.349 & 0.14 & 0.52 & 0.001\\
    & BERT & 98.31\% & 0.412 & 0.17 & 0.68 & 0.008\\
    & BERT Joint & 98.38\% & 0.432 & 0.17 & 0.63 & 0.004\\
    & RoBERTa & 98.51\% & 0.413 & 0.17 & 0.67 & 0.01\\
    & RoBERTa Joint & \textbf{98.57\%} & \textbf{0.426} & \textbf{0.18} & 0.63 & 0.008\\
    & GPT-2 & 98.44\% & 0.402 & 0.17 & \textbf{0.69} & 0.007\\ 
    & GPT-2 Joint & 98.47\% & 0.433 & 0.18 & 0.68 & 0.001\\\hline
    
  \end{tabular}
  \caption{Model performance results across different metrics. We use the integrated gradient to obtain the model's explanation. Model names with Joint keyword are trained with joint feature attribution loss. While, Token F1 and IOU F1 score denotes \emph{plausibility}, Comprehensiveness and Sufficiency denotes \emph{faithfulness}.}
\label{Table 2}
\end{table*}

\subsection{Experiment details}
We fine tune CNN \cite{kim2014convolutional}, LSTM \cite{hochreiter1997long}, bert-base-uncased (BERT) \cite{devlin-etal-2019-bert}, RoBERTa-base (RoBERTa) \cite{liu2019roberta}, GPT-2 \citep{radford2019language} models on both ATIS and SNIPS datasets. All the models are evaluated on both classification and explainability metrics. We calculate attribution scores using integrated gradients and LIME as it performs better for explainability properties than other post-hoc techniques \cite{atanasova-etal-2020-diagnostic}. For CNN, LSTM, BERT, and RoBERTa, we evaluate performance on explainability metrics with simple model training and models trained with joint attribution loss as stated in equation \ref{joint loss}. For training with joint loss, we guide the model to focus on silver annotated explanation signals. Attribution scores are calculated using integrated gradients for joint training. We maintain a constant number of 50 steps to calculate attribution scores using integrated gradients. Other hyperparameter details and model architecture details are provided in appendix \ref{sec:appendix}

\section{Evaluation}

\subsection{Evaluation metrics}
An accuracy metric is used to evaluate the model's predictability. We evaluate the explainability of all the models using \emph{plausibility} and \emph{faithfulness} metrics \cite{deyoung2019eraser}. \emph{Plausibility} compares the model's reasoning with the ground truth explanation. Here, we use a silver annotated explanation as ground truth. \emph{Faithfulness} serves as a metric for evaluating the degree to which the model's provided explanation aligns with its actual reasoning.

We report the \emph{Token F1 score} and \emph{IOU F1} scores for plausibility. Token F1 score measures the mean F1 score of a direct match between rationale derived from the model and ground truth explanation signals. IOU F1 score allows credit for partial matches. Intersection over Union (IOU) metric at the token level is defined as the measure of token overlap between two spans divided by the size of their combined token sets. It considers a prediction a match if it intersects with any ground truth rationales by more than 0.5. These partial matches are later used to compute the F1 score \cite{deyoung2019eraser}. We evaluate using the top 5 words with the highest attribution scores, as ground truth explanations typically consist of 4-5 words. For \emph{faithfulness} we evaluate models on \emph{Comprehensiveness} and \emph{Sufficiency} metrics \cite{deyoung2019eraser}

\noindent\textbf{Sufficiency} evaluates the extent to which the extracted rationales provide the necessary information for a model to make decisions. If model $m$ predicts class j for text sample $x_i$, then the probability of that prediction is denoted by $m(x_i)_j$. $r_i$ denotes the predicted rationale. Sufficiency is measured as Sufficiency = $m(x_i)_j - m(r_i)_j$.

\noindent\textbf{Comprehensiveness} measures the importance of the predicted rationale by evaluating the change in prediction when the rationale is removed from the text utterance. It is defined as:  
\begin{equation}
\text{Comprehensiveness} = m(x_i)_j - m(x_i \setminus r_i)_j
\end{equation}  
where \( m(x_i \setminus r_i)_j \) represents the probability of predicting class \( j \) when the rationale \( r_i \) is removed from the text sample \( x_i \).

\subsection{Results}
We find that models perform well on intent classification but do not perform well on explainability metrics \emph{plausibility} and \emph{faithfulness}. We observe that improving the reasoning of models using regularization helps improve the model's predictions for the classes with scarce data. Training with attribution loss results in a 3-4\% improvement in classification accuracy for both CNN and LSTM models and a slight improvement for the BERT and RoBERTa models on the ATIS dataset. (refer to table \ref{Table 2}). In particular, we observe that the model's prediction improves for classes with few training samples. For example, \emph{atis-meal, atis-ground-fare, and atis-flight-no} etc. We believe there is less improvement in BERT and RoBERTa because of the strong priors obtained during pre-training. Besides prediction, training models on silver annotated explanations improve the models' performance in explainability metrics. It increases the token F1 score by 3-4\% (Table \ref{Table 2} and Table \ref{Table 4}) for both datasets. In both datasets, BERT and RoBERTa are observed to make more human-like decisions. As a result, they have significantly higher token F1 scores than CNN and LSTM. For the ATIS dataset, performance in all explainability metrics drops significantly. A large number of fine-grained classes with subtle differences in meaning might be the reason for that. Even though CNN and LSTM struggle with the plausibility metric, they perform well in the sufficiency metric (table \ref{Table 2} and \ref{Table 4}). We also observe that joint training the performance of models on the faithfulness metric. The LIME explanation has better faithfulness than gradient-based explanations (Table \ref{Table 4}). This suggests the independence of all the explainability metrics as depicted by \cite{deyoung2019eraser} and \cite{atanasova-etal-2020-diagnostic}.

\section{Analysis}
We analyze the reasoning of models listed in table \ref{Table 2} using both LIME and integrated gradient explanations. It is observed that most of the classification is based on highly frequent words. Hence, models do not give good F1 scores for skewed datasets like ATIS (Table \ref{Table 3}). For example, in the ATIS dataset, text utterance \emph{At the Charlotte airport, how many different types of aircraft are there for us.} has gold intent \emph{atis-aircraft}. To explain gold intent, the \emph{aircraft} word must be present as an explanation. But BERT fine-tuned on ATIS dataset relies on word \emph{airport} as class \emph{atis-airport} has more training samples. Training it on silver annotations from our dataset improves the model's reasoning and helps to predict the correct class. Although the model focuses on the most frequent terms a lot, it understands the role of the main predicate well. As a result, the explanation of 70-80\% text utterances consists of the main predicate in words with top 5 attribution scores in all experiments. In the SNIPS dataset, all the models perform very well but struggle with explainability. For example, the class \emph{SearchScreeningEvent} model should focus on the words "movies," "Play" and "theater." However, it focuses on terms like names of the movies or actors. Due to abundant samples, deep learning models perform well on the SNIPS dataset but might not generalize well across multiple domains in scarce data scenarios.

\section{Conclusion, and Future work}
We introduce a novel silver annotation technique using the main predicate and arguments to annotate word-level explanation signals automatically for intent classification. Using this technique, we augment standard intent classification datasets ATIS and SNIPS. Hence, we contribute a first-of-its-kind benchmark dataset for explainability in intent classification consisting of 21k samples. We validate the quality of silver standard annotated explanations by detailed human evaluation. This approach can be adopted across diverse domains and intents. We evaluated several models on this dataset and found that models performing well on intent classification do not perform well in plausibility and faithfulness in explainability metrics. However, ingesting feature attribution priors improves the model's reasoning and overall performance.

In future work, one can attempt to improve the model's performance under scarce data scenarios by training it on explanation signals. It will be interesting to apply the concept of main predicate and arguments to other languages and build a multilingual benchmark dataset for explainability in intent classification.

\section*{Limitations}
We do not leverage ground truth intent to generate explanations. Although slots in original cases are not explanation signals in most cases, they provide some hint of gold intent in some samples. Identification of such cases can help to improve explanations. Since our technique uses knowledge and linguistic information like Parts of speech and dependency relations, it may not work best with multilingual settings because language structure differs for different languages.

\section*{Ethics Statement}
We use all the open-source datasets, and their use does not harm people's privacy using chatbots or any conversational AI platforms. All the models are trained on the same datasets; hence, it maintains people's privacy.

\noindent

\bibliographystyle{acl_natbib}
\bibliography{custom}

\begin{thebibliography}{31}
\expandafter\ifx\csname natexlab\endcsname\relax\def\natexlab#1{#1}\fi

\bibitem[{Atanasova et~al.(2020)Atanasova, Simonsen, Lioma, and Augenstein}]{atanasova-etal-2020-diagnostic}
Pepa Atanasova, Jakob~Grue Simonsen, Christina Lioma, and Isabelle Augenstein. 2020.
\newblock \href {https://doi.org/10.18653/v1/2020.emnlp-main.263} {A diagnostic study of explainability techniques for text classification}.
\newblock In \emph{Proceedings of the 2020 Conference on Empirical Methods in Natural Language Processing (EMNLP)}, pages 3256--3274, Online. Association for Computational Linguistics.

\bibitem[{Casanueva et~al.(2020)Casanueva, Tem{\v{c}}inas, Gerz, Henderson, and Vuli{\'c}}]{casanueva-etal-2020-efficient}
I{\~n}igo Casanueva, Tadas Tem{\v{c}}inas, Daniela Gerz, Matthew Henderson, and Ivan Vuli{\'c}. 2020.
\newblock \href {https://doi.org/10.18653/v1/2020.nlp4convai-1.5} {Efficient intent detection with dual sentence encoders}.
\newblock In \emph{Proceedings of the 2nd Workshop on Natural Language Processing for Conversational AI}, pages 38--45, Online. Association for Computational Linguistics.

\bibitem[{Chen et~al.(2019)Chen, Zhuo, and Wang}]{chen2019bert}
Qian Chen, Zhu Zhuo, and Wen Wang. 2019.
\newblock Bert for joint intent classification and slot filling.
\newblock \emph{arXiv preprint arXiv:1902.10909}.

\bibitem[{Coucke et~al.(2018)Coucke, Saade, Ball, Bluche, Caulier, Leroy, Doumouro, Gisselbrecht, Caltagirone, Lavril et~al.}]{coucke2018snips}
Alice Coucke, Alaa Saade, Adrien Ball, Th{\'e}odore Bluche, Alexandre Caulier, David Leroy, Cl{\'e}ment Doumouro, Thibault Gisselbrecht, Francesco Caltagirone, Thibaut Lavril, et~al. 2018.
\newblock Snips voice platform: an embedded spoken language understanding system for private-by-design voice interfaces.
\newblock \emph{arXiv preprint arXiv:1805.10190}.

\bibitem[{Devlin et~al.(2019)Devlin, Chang, Lee, and Toutanova}]{devlin-etal-2019-bert}
Jacob Devlin, Ming-Wei Chang, Kenton Lee, and Kristina Toutanova. 2019.
\newblock \href {https://doi.org/10.18653/v1/N19-1423} {{BERT}: Pre-training of deep bidirectional transformers for language understanding}.
\newblock In \emph{Proceedings of the 2019 Conference of the North {A}merican Chapter of the Association for Computational Linguistics: Human Language Technologies, Volume 1 (Long and Short Papers)}, pages 4171--4186, Minneapolis, Minnesota. Association for Computational Linguistics.

\bibitem[{DeYoung et~al.(2019)DeYoung, Jain, Rajani, Lehman, Xiong, Socher, and Wallace}]{deyoung2019eraser}
Jay DeYoung, Sarthak Jain, Nazneen~Fatema Rajani, Eric Lehman, Caiming Xiong, Richard Socher, and Byron~C Wallace. 2019.
\newblock Eraser: A benchmark to evaluate rationalized nlp models.
\newblock \emph{arXiv preprint arXiv:1911.03429}.

\bibitem[{Elazar et~al.(2021)Elazar, Ravfogel, Jacovi, and Goldberg}]{elazar-etal-2021-amnesic}
Yanai Elazar, Shauli Ravfogel, Alon Jacovi, and Yoav Goldberg. 2021.
\newblock \href {https://doi.org/10.1162/tacl_a_00359} {Amnesic probing: Behavioral explanation with amnesic counterfactuals}.
\newblock \emph{Transactions of the Association for Computational Linguistics}, 9:160--175.

\bibitem[{Gunaratna et~al.(2022)Gunaratna, Srinivasan, Yerukola, and Jin}]{gunaratna2022explainable}
Kalpa Gunaratna, Vijay Srinivasan, Akhila Yerukola, and Hongxia Jin. 2022.
\newblock Explainable slot type attentions to improve joint intent detection and slot filling.
\newblock \emph{arXiv preprint arXiv:2210.10227}.

\bibitem[{Hayati et~al.(2021)Hayati, Kang, and Ungar}]{hayati-etal-2021-bert}
Shirley~Anugrah Hayati, Dongyeop Kang, and Lyle Ungar. 2021.
\newblock \href {https://doi.org/10.18653/v1/2021.emnlp-main.510} {Does {BERT} learn as humans perceive? understanding linguistic styles through lexica}.
\newblock In \emph{Proceedings of the 2021 Conference on Empirical Methods in Natural Language Processing}, pages 6323--6331, Online and Punta Cana, Dominican Republic. Association for Computational Linguistics.

\bibitem[{Hemphill et~al.(1990)Hemphill, Godfrey, and Doddington}]{hemphill-etal-1990-atis}
Charles~T. Hemphill, John~J. Godfrey, and George~R. Doddington. 1990.
\newblock \href {https://aclanthology.org/H90-1021} {The {ATIS} spoken language systems pilot corpus}.
\newblock In \emph{Speech and Natural Language: Proceedings of a Workshop Held at Hidden Valley, {P}ennsylvania, June 24-27,1990}.

\bibitem[{Hochreiter and Schmidhuber(1997)}]{hochreiter1997long}
Sepp Hochreiter and J{\"u}rgen Schmidhuber. 1997.
\newblock Long short-term memory.
\newblock \emph{Neural computation}, 9(8):1735--1780.

\bibitem[{Jayaram and Allaway(2021)}]{jayaram-allaway-2021-human}
Sahil Jayaram and Emily Allaway. 2021.
\newblock \href {https://doi.org/10.18653/v1/2021.emnlp-main.450} {Human rationales as attribution priors for explainable stance detection}.
\newblock In \emph{Proceedings of the 2021 Conference on Empirical Methods in Natural Language Processing}, pages 5540--5554, Online and Punta Cana, Dominican Republic. Association for Computational Linguistics.

\bibitem[{Joshi et~al.(2021)Joshi, Chatterjee, and Ekbal}]{joshi2021towards}
Ratnesh Joshi, Arindam Chatterjee, and Asif Ekbal. 2021.
\newblock Towards explainable dialogue system: Explaining intent classification using saliency techniques.
\newblock In \emph{Proceedings of the 18th International Conference on Natural Language Processing (ICON)}, pages 120--127.

\bibitem[{Kim(2014)}]{kim2014convolutional}
Yoon Kim. 2014.
\newblock Convolutional neural networks for sentence classification.
\newblock \emph{arXiv preprint arXiv:1408.5882}.

\bibitem[{Landis and Koch(1977)}]{landis1977measurement}
J~Richard Landis and Gary~G Koch. 1977.
\newblock The measurement of observer agreement for categorical data.
\newblock \emph{biometrics}, pages 159--174.

\bibitem[{Li et~al.(2022)Li, Chen, Shen, Chen, Zhang, Li, Wang, Qian, Peng, Mao et~al.}]{li2022explanations}
Shiyang Li, Jianshu Chen, Yelong Shen, Zhiyu Chen, Xinlu Zhang, Zekun Li, Hong Wang, Jing Qian, Baolin Peng, Yi~Mao, et~al. 2022.
\newblock Explanations from large language models make small reasoners better.
\newblock \emph{arXiv preprint arXiv:2210.06726}.

\bibitem[{Liu and Avci(2019)}]{liu2019incorporating}
Frederick Liu and Besim Avci. 2019.
\newblock Incorporating priors with feature attribution on text classification.
\newblock \emph{arXiv preprint arXiv:1906.08286}.

\bibitem[{Liu et~al.(2019)Liu, Ott, Goyal, Du, Joshi, Chen, Levy, Lewis, Zettlemoyer, and Stoyanov}]{liu2019roberta}
Yinhan Liu, Myle Ott, Naman Goyal, Jingfei Du, Mandar Joshi, Danqi Chen, Omer Levy, Mike Lewis, Luke Zettlemoyer, and Veselin Stoyanov. 2019.
\newblock Roberta: A robustly optimized bert pretraining approach.
\newblock \emph{arXiv preprint arXiv:1907.11692}.

\bibitem[{Mathew et~al.(2021)Mathew, Saha, Yimam, Biemann, Goyal, and Mukherjee}]{mathew2021hatexplain}
Binny Mathew, Punyajoy Saha, Seid~Muhie Yimam, Chris Biemann, Pawan Goyal, and Animesh Mukherjee. 2021.
\newblock Hatexplain: A benchmark dataset for explainable hate speech detection.
\newblock In \emph{Proceedings of the AAAI conference on artificial intelligence}, volume~35, pages 14867--14875.

\bibitem[{Montavon et~al.(2019)Montavon, Binder, Lapuschkin, Samek, and M{\"u}ller}]{montavon2019layer}
Gr{\'e}goire Montavon, Alexander Binder, Sebastian Lapuschkin, Wojciech Samek, and Klaus-Robert M{\"u}ller. 2019.
\newblock Layer-wise relevance propagation: an overview.
\newblock \emph{Explainable AI: interpreting, explaining and visualizing deep learning}, pages 193--209.

\bibitem[{Qin et~al.(2021)Qin, Liu, Che, Kang, Zhao, and Liu}]{qin2021co}
Libo Qin, Tailu Liu, Wanxiang Che, Bingbing Kang, Sendong Zhao, and Ting Liu. 2021.
\newblock A co-interactive transformer for joint slot filling and intent detection.
\newblock In \emph{ICASSP 2021-2021 IEEE International Conference on Acoustics, Speech and Signal Processing (ICASSP)}, pages 8193--8197. IEEE.

\bibitem[{Radford et~al.(2019)Radford, Wu, Child, Luan, Amodei, Sutskever et~al.}]{radford2019language}
Alec Radford, Jeffrey Wu, Rewon Child, David Luan, Dario Amodei, Ilya Sutskever, et~al. 2019.
\newblock Language models are unsupervised multitask learners.
\newblock \emph{OpenAI blog}, 1(8):9.

\bibitem[{Raymond and Riccardi(2007)}]{raymond2007generative}
Christian Raymond and Giuseppe Riccardi. 2007.
\newblock Generative and discriminative algorithms for spoken language understanding.
\newblock In \emph{Interspeech 2007-8th Annual Conference of the International Speech Communication Association}.

\bibitem[{Ribeiro et~al.(2016)Ribeiro, Singh, and Guestrin}]{ribeiro2016should}
Marco~Tulio Ribeiro, Sameer Singh, and Carlos Guestrin. 2016.
\newblock " why should i trust you?" explaining the predictions of any classifier.
\newblock In \emph{Proceedings of the 22nd ACM SIGKDD international conference on knowledge discovery and data mining}, pages 1135--1144.

\bibitem[{Ribeiro et~al.(2018)Ribeiro, Singh, and Guestrin}]{ribeiro2018anchors}
Marco~Tulio Ribeiro, Sameer Singh, and Carlos Guestrin. 2018.
\newblock Anchors: High-precision model-agnostic explanations.
\newblock In \emph{Proceedings of the AAAI conference on artificial intelligence}, volume~32.

\bibitem[{Sundararajan et~al.(2017)Sundararajan, Taly, and Yan}]{sundararajan2017axiomatic}
Mukund Sundararajan, Ankur Taly, and Qiqi Yan. 2017.
\newblock Axiomatic attribution for deep networks.
\newblock In \emph{International conference on machine learning}, pages 3319--3328. PMLR.

\bibitem[{Wang et~al.(2016)Wang, Huang, Zhu, and Zhao}]{wang-etal-2016-attention}
Yequan Wang, Minlie Huang, Xiaoyan Zhu, and Li~Zhao. 2016.
\newblock \href {https://doi.org/10.18653/v1/D16-1058} {Attention-based {LSTM} for aspect-level sentiment classification}.
\newblock In \emph{Proceedings of the 2016 Conference on Empirical Methods in Natural Language Processing}, pages 606--615, Austin, Texas. Association for Computational Linguistics.

\bibitem[{Wu and Ong(2021)}]{wu2021explaining}
Zhengxuan Wu and Desmond~C Ong. 2021.
\newblock On explaining your explanations of bert: An empirical study with sequence classification.
\newblock \emph{arXiv preprint arXiv:2101.00196}.

\bibitem[{Xu et~al.(2015)Xu, Ba, Kiros, Cho, Courville, Salakhudinov, Zemel, and Bengio}]{pmlr-v37-xuc15}
Kelvin Xu, Jimmy Ba, Ryan Kiros, Kyunghyun Cho, Aaron Courville, Ruslan Salakhudinov, Rich Zemel, and Yoshua Bengio. 2015.
\newblock \href {https://proceedings.mlr.press/v37/xuc15.html} {Show, attend and tell: Neural image caption generation with visual attention}.
\newblock In \emph{Proceedings of the 32nd International Conference on Machine Learning}, volume~37 of \emph{Proceedings of Machine Learning Research}, pages 2048--2057, Lille, France. PMLR.

\bibitem[{Ye and Durrett(2022)}]{ye2022unreliability}
Xi~Ye and Greg Durrett. 2022.
\newblock The unreliability of explanations in few-shot prompting for textual reasoning.
\newblock \emph{Advances in neural information processing systems}, 35:30378--30392.

\bibitem[{Zhong et~al.(2019)Zhong, Shao, and McKeown}]{zhong2019fine}
Ruiqi Zhong, Steven Shao, and Kathleen McKeown. 2019.
\newblock Fine-grained sentiment analysis with faithful attention.
\newblock \emph{arXiv preprint arXiv:1908.06870}.

\end{thebibliography}

\appendix

\section{Appendix}
\label{sec:appendix}

\subsection{Model details}

\subsubsection{CNN}
We build a simple text classifier CNN model. The architecture consists of 256 filters, 2,3 and 4-gram. The network takes into input sequence length of 80. Following convolution layers, architecture consists of max-pooling and softmax function layers. The network performs well with the SNIPS dataset and has decent performance for the ATIS dataset. In addition to the basic network, we ingest feature attribution during training to provide supervision over rationale. These feature attributions are calculated using an integrated gradient technique.

\subsubsection{LSTM}
For the LSTM model, a sequence of length 80 is passed as embeddings. The network consists of 2 fully connected layers of neurons with 32 and 32 neurons each. We also added a dropout layer to avoid overfitting. Like the CNN model, LSTM is also trained on joint attribution loss.

\subsection{Hyperparameters details}
All the training has been conducted on NVIDIA A100-SXM4-80GB GPU. We use a batch size of 32 for all the experiments. We use the cross-entropy loss function and the Adam optimizer for training. For the simple fine-tuning, the value of $\lambda$ is kept at 0. To fix the value of hyperparameter $\lambda$, we extensively search the range values $(1,10^{10})$ and select the values for which the best results are obtained. For joint attribution training of CNN and LSTM, $\lambda$ is set to $ 10^{5} $ and $ 10^{4} $ for the SNIPS dataset. For the ATIS dataset, $\lambda$ values are set as $ 10^{6} $ for both CNN and LSTM models. The initial learning rate for BERT is 4e-5, and 0.001 for both CNN and LSTM. BERT was trained for 2, CNN for 10, and LSTM for five epochs for both datasets. All the training details for the Roberta and GPT-2 model are the same as those for BERT.

\begin{table*}[t]\centering
  \begin{tabular}{c c c c c c c}
    Dataset & Model & Token F1 & IOU F1 & Comprehensiveness & Sufficiency\\ \hline
    \multirow{8}{*}{\textbf{ATIS}}& CNN & 0.196 & 0.03 &0.42 & \textbf{-0.072}\\
    & CNN Joint & 0.228 & 0.03 & 0.38 & -0.37\\
    & LSTM & 0.182 & 0.05 & 0.44 & -0.06\\
    & LSTM Joint & 0.203 & 0.05 & 0.47 & -0.09\\
    & BERT & 0.26 & 0.09 & \textbf{0.53} & 0.09\\
    & BERT Joint & 0.298 & 0.1 & 0.49 & 0.05\\
    & RoBERTa & 0.253 & 0.09 & 0.51 & 0.1\\
    & RoBERTa Joint & \textbf{0.302} & \textbf{0.11} & 0.49 & 0.06\\ \hline
    \multirow{8}{*}{\textbf{SNIPS}}& CNN & 0.33 & 0.14 & 0.62 & \textbf{-0.081}\\
    & CNN Joint & 0.36 & 0.14 & 0.55 & -0.085\\
    & LSTM & 0.33 & 0.14 & 0.55 & 0.001\\
    & LSTM Joint & 0.34 & 0.14 & 0.57 & -0.42\\
    & BERT & 0.37 & 0.16 & \textbf{0.72} & 0.005\\
    & BERT Joint & \textbf{0.4} & \textbf{0.17} & 0.69 & 0.002\\
    & RoBERTa & 0.36 & 0.16 & 0.7 & 0.2\\
    & RoBERTa Joint & 0.39 & 0.17 & 0.63 & 0.03\\ \hline
    
  \end{tabular}
  \caption{Model performance results across different metrics. Here, LIME is used to obtain the model's explanation. Model names with Joint keyword are trained with joint feature attribution loss}
\label{Table 4}
\end{table*}

\begin{table*}[t]
  \begin{tabular}{c c c c c}
    Class names & precision & recall & F1 score & samples\\ \hline
    atis-abbreviation & 0.94 & 1.0 & 0.97 & 33 \\
    atis-aircraft & 0.8 & 0.89 & 0.84 & 9 \\
    atis-airfare & 0.77 & 1.00 & 0.87 & 48 \\
    atis-airline & 0.9 & 1.0 & 0.95 & 38 \\
    atis-airport & 0.95 & 1.0 & 0.97 & 18 \\
    atis-capacity & 1.00 & 0.52 & 0.69 & 21 \\
    atis-city & 1.00 & 0.67 & 0.8 & 6 \\
    atis-distance & 0.77 & 1.00 & 0.87 & 10 \\
    atis-flight & 0.99 & 0.99 & 0.99 & 632 \\
    atis-flight-no & 1.00 & 0.38 & 0.55 & 8 \\
    atis-flight-time & 0.5 & 1.00 & 0.67 & 1 \\
    atis-ground-fare & 1.00 & 0.57 & 0.73 & 7 \\
    atis-ground-service & 0.92 & 1.00 & 0.96 & 36 \\
    atis-meal & 1.00 & 0.33 & 0.5 & 6 \\
    atis-quantity & 0.25 & 1.00 & 0.4 & 3
  \end{tabular}
  \caption{BERT model class-wise performance when fine-tuned on ATIS dataset. Column samples denote the number of samples for that class.}
\label{Table 3}
\end{table*}

\end{document}